# Exploring the Efficacy of Group-Normalization in Deep Learning Models for Alzheimer's Disease Classification

Dr. Gousia Habib[1*]
*Postdoctoral Researcher, EE*
*IIT Delhi, Hauz Khas, India.*

Dr. Ishfaq Ahmed Malik[2]
*Shoolini University Solan HP, India*

Dr. Jameel Ahmad[3]
*MANNU University Hyderabad, India*

Imtiaz Ahmed[4]
*NIT Srinagar, J&K, India*

Dr. Shaima Qureshi[5]
*Assistant Professor, CSE*
*NIT Srinagar, J&K, India*

**Abstract**

*Batch Normalization (B.N.) is an important approach to advancing deep learning since it allows multiple networks to train simultaneously. A problem arises when normalizing along the batch dimension because B.N.'s error increases significantly as batch size shrinks because batch statistics estimates are inaccurate. As a result, computer vision tasks like detection, segmentation, and video, which require tiny batches based on memory consumption, aren't suitable for using B.N. for larger model training and feature transfer. Here, we explore Group Normalization (G.N.) as an easy alternative to using B.N. A GN is a channel normalization method in which each group is divided into different channels, and the corresponding mean and variance are calculated for each group. G.N.'s computations are accurate across a wide range of batch sizes and are independent of batch size. When trained using a large ImageNet database on ResNet-50, GN achieves a very low error rate of 10.6% compared to B.N. when a smaller batch size of only 2 is used. For usual batch sizes, the performance of G.N. is comparable to that of B.N., but at the same time, it outperforms other normalization techniques. Implementing G.N. as a direct alternative to B.N. to combat the serious challenges faced by the B.N. in deep learning models with comparable or improved classification accuracy. Additionally, G.N. can be naturally transferred from the pre-training to the fine-tuning phase. Our work implements ResNet-50 on a medical image dataset for Alzheimer's disease classification. Compared with existing state-of-the-art techniques, the proposed model outperforms them in terms of accuracy (ACC), sensitivity (SEN), specificity (SPE), and Mathew's correlation coefficient (MCC). The proposed methods achieve an accuracy of 95.5 % with a minimum training loss, thus outperforming all the state-of-the-art techniques.*

**Keywords:** Group Normalization (GN), Weight Normalization (WN), ResNet-50, Batch Normalization (BN), Instance Normalization (IN).

## 1. Introduction

In machine learning, normalization refers to scaling and transforming input data to improve performance and stability. Normalization adjusts the range and distribution of input features so that they have similar statistical properties, thus making them easier for the model to learn. A major advantage of normalization is that it prevents internal covariate shifts in machine learning models. The internal covariate shift occurs when the distribution of input features changes as the data flows through different layers. As the model's weights are updated, the activations of each layer can change due to different updates in the model's weights. An internal covariate shift can make it difficult for the model to reach an optimal solution, and it can lead to poor performance and long training times. As a result of normalizing the input features, we can reduce the impact of internal covariate shift by ensuring that the input distributions are stable and consistent at different layers. Several types of normalization techniques can be used in machine learning, including BN, LN, IN, and GN. Each of these methods has its own advantages and disadvantages, and the choice of normalization method will depend on the specific model and problem.

Machine learning models can also benefit from normalization in order to reduce overfitting. The overfitting process occurs when models become too complex and begin to fit to noise or random fluctuations in training data, rather than underlying patterns. Overfitting can be prevented by normalizing the weights in the model in order to reduce their magnitude. Input features are normalized so that the range and distribution of the input data are compressed, which means that the weights in the model do not need to be as large to capture the same amount of data. In this way, overfitting can be minimized and the model can be more robust. As well, normalization introduces noise and randomness into the training process, which helps to regularize the model. A model can improve its ability to generalize and become less dependent on any particular feature or pattern in the data if it isn't too dependent on one feature or pattern in the data.

Overall, normalization is one of the most important machine learning techniques that can help to improve a model's performance, stability, and generalization. The adjustments to the range and distribution of input features by normalization can mitigate the impact of internal covariate shifts and prevent overfitting, which are common challenges for deep learning and other complex models. In addition to reducing internal covariate shifts and preventing overfitting, it makes models more robust to variations in input data due to its ability to reduce internal covariate shifts. Moreover, normalization can improve the interpretability of a model by improving the understanding of how different features contribute to the final result. In general, normalization plays a key role in building high-quality machine learning models that can solve complex problems across a broad range of domains.

As a result, understanding and applying normalization techniques is essential to building effective models in the field of data science and machine learning. The most popular types of normalization techniques used in machine learning models are discussed as:

### 1.1. Batch Normalization

B.N is a methodology for normalizing network activations throughout a definite-size mini-batch. It estimates the mean and variance of every individual feature in the mini-batch. The resulting feature is calculated by division with a standard deviation of the mini-batch followed by subtraction from the mean. The application of batch normalization makes learning more efficient, and it may also be used as a regularization to prevent overfitting issues. B.N. layer can be added to any sequential model to provide normalized inputs or outputs. B.N. can be used in various places throughout the model's layers. Normally positioned just after the sequential model's definition and before the convolution and pooling layers. Batch normalization is a two-step process; input normalization is the first step, and rescaling and offsetting are performed. Some advantages of batch normalization are listed: It increases the Training Time, minimizes the internal covariate shift (ICS) and speeds up deep neural network training. This method lowers the reliance of gradients on parameter scales or initial values, resulting in greater learning rates and unhazardous convergence. It allows saturating nonlinearities to be used since it prevents the network from becoming stuck in saturated modes.

**1.2. Layer Normalization**

Layer normalization is a technique for speeding up the training of neural network models. Unlike batch normalization, this approach estimates normalization statistics directly from the summed inputs to hidden layer neurons. To overcome the major drawback of B.N. of actually being reliable on the small batch sizes, a new normalization method came into existence known as L.N. The following points can depict the benefits of L.N: L.N. is simple to implement on Recurrent Neural Networks (RNNs) by estimating the normalization statistics independently at each time step. This method is powerful in recurrent neural networks for stabilizing hidden state dynamics.

The average size of enumerated inputs to the recurrent units tends to expand or recoil at every step in a normal RNN, resulting in explosive or vanishing gradients. A layer normalized RNN shows invariance to rescaling all enumerated inputs to a layer due to the normalization parameters, leading to substantially more stable hidden-to-hidden dynamics.

**1.3. Weight Normalization**

By decoupling the length and direction of the weight vectors, weight normalization (W.N.) re-parametrizes the weight vectors of deep neural networks. In simple terms, W.N. is considered a method that increases the ability of neural networks to optimize their weights. The benefits of W.N. are listed as follows: It improves the optimization problem's conditioning and speeds up stochastic gradient descent convergence. It works well for recurrent models like LSTMs, reinforcement learning and deep generative models.

**1.4. Instance (or Contrast) Normalization**

L.N and normalization by instances (IN) have many similarities. Nevertheless, IN is standardized over each channel in each training example. However, L.N. standardizes overall input characteristics in a training sample. In contrast to B.N., the IN layer is also implemented during testing (due to the non-dependence of the mini-batch). It has been proposed as an alternative method for batch normalization in GANs and is utilized in style transfer applications. Some of the advantages of the IN are listed as follows: It makes a model's learning process easier. It can be best employed at the test time.

### 1.5. Switchable Normalization

This method employs a weighted average of numerous mean and variance statistics derived from B.N., IN, and L.N. According to recent research trends, Switch Normalization (S.N) could potentially outperform batch normalization on various computer vision tasks of image and object detection and categorization. For estimation of statistical quantities (means and variances), A channel, a layer, and a mini-batch makeup S.N.'s three fields. As S.N. learns the importance weights for each, it alternates between them in an end-to-end manner. It has a lot of wonderful qualities. It adapts to a variety of network designs and workloads, for starters. Secondly, it can handle a broader range of batch sizes, retaining efficiency even for smaller Minibatches (e.g.,2 images/GPU) is provided. A summary of various normalization techniques discussed in succeeding sections is given in Table 1. The entire paper is divided into 6 sections. Section 1 provides a brief introduction of various normalization techniques used and the major challenges posed by them. This section motivates us to propose a novel group normalization technique. Section 2 discusses major challenges faced by the B.N. and the motivation behind implementing G.N. in this research article. Section 3 gives the main contribution of the work. Section 4 gives related work normalization techniques currently being used. Section 5 provides insights into the proposed algorithm and model for evaluating the performance of the novel algorithm. Section 6 discusses the related results and results obtained after implementing the proposed techniques. Finally, the paper is concluded at the end, and some of the limitations and future scope of the work is provided.

### 2. Problem Statement and Motivation behind Group Normalization

The B.N. calculates the batch statistics (mini-batch mean and variance) in every training iteration, requiring larger batch sizes to approximate the population mean and variance from the mini-batch. Due to their high input resolution (often as big as 1024x2084), B.N. networks make it harder to train for applications such as object detection, semantic segmentation, etc., since they require training with large batch sizes. It is also important to note that the B.N. layer does not calculate the mean and variance from the test data mini-batch during test (or inference) time. But instead, it uses the fixed mean and variance calculated from the training data. Due to this, B.N. must be used with caution, and a great deal of complexity is introduced. We need to fix the issue, such as when each data sample is highly memory intensive, as in the case of video or high-resolution images. Training larger neural networks may consume GPU processors, leaving little memory for processing data. It is, therefore, necessary to have alternatives to B.N. that work well with small batch sizes. As one of the most recent normalization methods, G.N does not exploit the batch dimension and therefore does not depend on the batch size.

The GN process is similar to LN since it is applied along the feature direction; however, instead of normalizing all the features at once, it divides them into certain groups and normalizes them separately. The parameter Num groups can be tuned as a hyperparameter to make Group normalization perform better than any other normalization technique employed for deep learning models.

This paper introduces a group normalization technique as a straightforward alternative to BN. Normalizing along the batch dimension is not feasible as B.N.'s error grows significantly as the batch size shrinks due to poor batch statistics estimates. So, introducing the group normalization method is the most effective normalizing layer without relying on batch dimensions. The resnet -50 model is used for validating the performance of the G.N. algorithm and conventional B.N. algorithm. Both conventional well-proposed algorithms are tested on Alzheimer's dataset, and corresponding performance measures are observed and recorded. Then the comparative analysis between state-of-the-art and novel techniques is observed and compared in terms of classification accuracy, training error, validation error, MCC, SPE, and SEN.

Table 1. Summary of Normalization Techniques with Research Gaps

| Normalization Technique | Mathematical Formulation | Graphical Visualization | Research Gaps |
|---|---|---|---|
| Batch Normalization | $\mu_B = \frac{1}{n}\sum_{i=1}^{n} y_i$ <br> // minibatch mean <br><br> $\sigma_B^2 = \frac{1}{n}\sum_{i=1}^{n}(y_i - \mu_B)^2$    (1) <br> // minibatch variance | 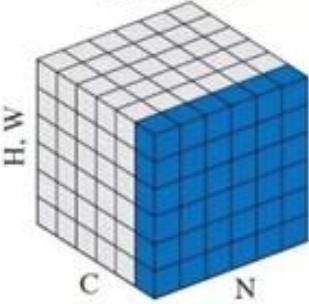 | 1. Lacking proper batch statistical estimation, B.N.'s error quantity increases more rapidly with a decrease in batch size. <br><br> 2. This restricts the use of B.N. for bigger model training and feature transfer to computer vision tasks, which require tiny batches bound by memory constraints. |
| Layer Normalization | $\mu_k = \frac{1}{n}\sum_{j=1}^{n} y_{k,j}$ <br> // minibatch mean <br><br> $\sigma_k^2 = \frac{1}{n}\sum_{j=1}^{n}(y_{k,j} - \mu_k)^2$ <br> // minibatch variance <br><br> $y_{k,j} = \frac{y_{k,j} - \mu_j}{\sqrt{\sigma_k^2 + \epsilon}}$    (2) | 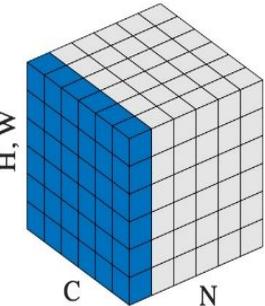 | 1. The parameters of the layer norm consisting of bias and gain sometimes increase the risk of overfitting the model and do not perform well. |
| Weight Normalization | $w = \dfrac{g}{\|v\|} v \quad (3)$<br><br>Reparametrizes the weight W | 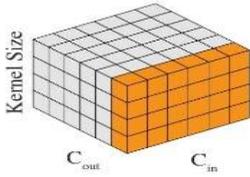 | 1. Training is performed in a similar way to batch normalization. Still, it provides significantly less stable training than batch normalization, which restricts its implementation in many applications. |
| Instance Normalization | $\mu_{qi} = \dfrac{1}{hw} \sum_{p=1}^{w} \sum_{r=1}^{h} z_{tipr}$<br><br>$\sigma_{qi}^{2} = \dfrac{1}{hw} \sum_{p=1}^{w} \sum_{r=1}^{h} \left( z_{tipr} - n\mu_{qi} \right)^{2}$<br><br>(4) | 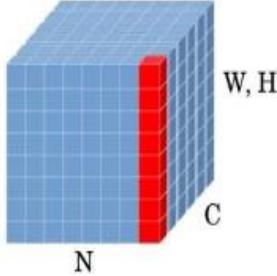 | 1. The technique is mainly designed only for style transfer. It is restricted to addressing the problem of the network being agnostic to the contrast of the original image. |
| Switchable Normalization | $\hat{y}_{mqlm} = \gamma \dfrac{y_{mqlm} - \sum_{k \in \Omega} w_{k}^{T}}{\sum_{k \in \Omega} w_{k}^{T} \sigma_{k}^{2}} + \beta$<br><br>(5)<br><br>$\Omega$ set of statistics estimated in different ways | 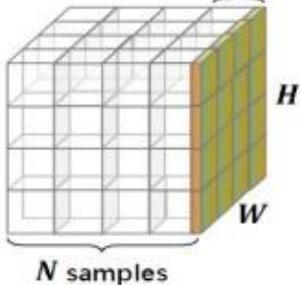 | 1. Sometimes, it gives low classification scores compared to its hybrid technique, switchable-shake normalization (SSN) [1]. |

## 4. Related Work

Overfitting and extended training times are two major issues in multilayer perceptron's (MLPs) and Deep Neural Networks (DNNs). Dropout and B.N. are two well-known methods for addressing these issues. Even though both techniques have overlapping design principles, several study findings have proven that they offer distinct advantages in improving deep learning.

**(Christian Garbin, Xing Quan Zhu X et al. in 2020)** Investigated the effects of dropout and B.N. on MLPs and convolutional neural networks (CNNs) separately and jointly. In this paper, the authors showed that B.N. enhanced accuracy unhazardous. B.N. should be one of the first stages in optimizing a CNN because it may be added without causing large structural changes to the network architecture. As the B.N. [2] suggested, increasing the learning rate improves the accuracy rate by 2% to 3%. This is one of the basic steps that must be taken before moving on to more difficult optimizations. Batch Normalization (Batch Norm) has become the standard component for training stability in current neural networks. For normalizing features over the batch dimension, Batch Norm uses centering and scaling procedures and mean and variance statistics [3]**.**

**(Shang-Hua Gao, Qi Han et la in 2021)** To improve instance-specific representations while maintaining the powerful uses of Batch Norm, the authors in this paper offers Representative Batch Normalization (RBN) with a very simple and efficient feature calibration technique. The centering calibration increases the strength of useful features while decreasing the strength of noisy features [4].

**(Ruibin Xiong, Yunchang Yang et al. in 2020)** The authors in this work investigated why the learning rate warm-up stage is significant in Transformer training and demonstrated that the L.N. position matters. They showed that the predicted gradients of the parameters of the output layer are significant at initiation in the original Transformer, placing the L.N. outside the residual blocks. When employing a high learning rate. It results in unstable training. It was also shown that the Transformer that placed L.N. within residual blocks could be trained despite the warm-up stage and even showed convergence more quickly [5].

**(Fenglin Liu, Xuancheng Ren et al 2020)** In this paper, the authors investigated how to skip connection and L.N. may be used to create a unique skip connection architecture that can integrate adequate skip information at the output le l for providing a solution to the gradient explosion problem in practice. As per empirical analysis, the authors observed that expansion of skip information helps aid model optimization, which is impeded in real-time by Gradient miscreation. L.N. performs a better job of minimizing such a problem than B.N. [6].

**(Yongcheng Jing, Xiao Liu et al. in 2020)** This study presented a new dynamic IN layer (DIN) for competent style transfers. The suggested DIN eliminates the problem of redundancy and sharable encoders in earlier techniques by allowing an extensive style encoder to encode rich style patterns and a lightweight content encoder to increase efficiency. Experiments show that the suggested method outperforms the current state of the art, particularly when transferring difficult style patterns while maintaining a low computational cost [7].

**(Liangyu Chen, Xin Lu et al in 2021)** The normalization techniques are particularly employed for image restoration jobs in this work. In particular, we bring IN into a residual block and create the Half Instance Normalization Block (HIN Block), which is both effective and efficient. We use Instance Normalization for half of the intermediate features in the HIN Block while keeping the content information. We also propose HINet, a multi-stage network based on HIN Block. We employ feature fusion and an attention-guided map [8] between stages to improve the multi-scale feature expression and facilitate the flow of information. On various image restoration tasks, our suggested HINet outperforms the SOTA. We also won first place in the NTIRE 2021 Image De-Blurring Challenge - Track2. JPEG Artifacts [9] utilizing HINet [10].

**(Dawa C. Lepcha, Bhawna Goyal et al in 2020)** Developed a new method for fusing the pictures in this paper by using a bilateral cross filter for grey-level similarities and geometric closeness of neighboring pixels without smoothing edges. Then, for scale-aware operation, the resultant images acquired after subtraction from the cross bilateral filter image output from the original images are filtered through the rolling guidance filter. It removes small-scale structures, retains the rest of the image information, and successfully recovers the edges of resultant images. Finally, the images were fused using a weighted calculated technique and W.N. The results have been qualitatively and quantitatively validated, and a comparison is made with other existing state-of-the-art approaches. It was revealed that the suggested strategy outperforms existing picture fusion algorithms [11].

**(Xu Yuan, Xiangjun Shen et al in 2021)** Based on structural learning in deep neural networks, we introduced an enhanced deep structural weight normalization (DSWN) technique in this research. The proposed method was provided in conjunction with network structural data in the form of orthogonal weight normalization (OWN). The efficacy of DSWN-NM and DSWN-SM algorithms is verified by experimental results obtained from various computer vision datasets. A comparative analysis was made with alternative existing normalizations and sparsity methods [12].

**(Ping Luo, Ruimao Zhang et al. in 2019)** Proposed Switchable Normalization (S.N.) learns how to choose various normalization methods for deeper neural networks to solve a learning-to-normalize challenge. For estimating statistical quantities such as means and variances, S.N. uses three fields: a channel, a layer, and Minibatches. S.N. alternates between them in an end-to-end manner by learning their importance weights. According to our findings, S.N. provides the efficient feature for balanced learning and generalization while DNN training. Investigative research about S.N. aids in the comprehension of various normalizing techniques. [13].

**(Dhiraj Neupane, Yunsu Kim et al in 2021)** In this study, the authors use the benefits of various signal processing techniques. Such as continuous wavelet transforms for feature generation from a raw time-domain CWRU bearing database, a broadly employed dataset for detecting and diagnosing automatic machine bearing failures and is accepted as a basic model validation model. It is implemented using a switchable normalization-based convolutional neural network. (SN-CNN) for feature extracting purposes, detection and categorizing the normal (healthy) and faecal (defective) bearings. We used four different sets of 48k drive-end bearing data in our experiment. When scalogram images are input, the findings reveal that the suggested system achieves state-of-the-art accuracy [14].

**(Kaixiong Zhou, Xiao Huang et al. in 2020)** The group distance ratio and instance information gain are two over-smoothing measures based on the graph structures we present in this study. We introduce a novel normalization layer, DGN, to increase model performance against over smoothing by examining GNN models through the prism of these two metrics. It separates node representations of distinct classes by normalizing each set of related nodes independently. Experiments on real-world classification tasks revealed that DGN significantly delayed performance degradation by addressing the issue of over-smoothing [15].

**(Xiao-Yun Zhou, Jiacheng Sun et al, in 2020)** Batch group normalization (BGN) is presented with good performance, stability, and generalizability while avoiding additional trainable parameters, information from many layers or iterations, or additional computation. BGN makes it easier to calculate noisy/confused statistics in B.N. by adaptively inserting feature instances from the grouped (channel, height, and width) dimensions and controlling the size of divided feature groups with a hyper-parameter G [16].

**(Agus-gunawan, Xu Yin et al, in 2022)** The study explains G.N.'s working principles and some factors behind G.N.'s inferior performance against B.N. As part of the analysis, the authors use two numerical metrics, loss landscape and gradient predictiveness, to describe and evaluate the normalization effects in the training process. Their analysis is complemented by the design of a new normalization layer built on top of the G.N. by incorporating B.N.'s mechanism into the G.N. Their study is the first to analyze the issues associated with G.N. and make improvements to it [17]. The proposed normalization layer (G.N. layer) helps to ease network training of convolutional neural networks [18],[19] (measured by loss landscape and gradient predictiveness). Compared with the original G.N., the proposed layer's loss landscape and gradient predictiveness have a smaller range of values. This study shows that the proposed layer provides better training stability.

**(Gilbert Chandra, Srinivasulu Reddy et al. in 2023)** The improved performance of GN is evident from [20]. Within a batch, BN normalizes features based on means and variances. As the GN method normalizes the channels by dividing them into groups and computing the mean and variance within each group. GN calculates mean and variance along the Height and Width (H, W) axes and along groups of channels (C). The performance of GN is good compared to BN when batch sizes are typical, and it stands out from the other variations of normalization as well. A GN is naturally capable of being transformed from free training to fine-tuning, and its computation is independent of the size of the batch. GN is utilized in the proposed 2D-CNN [20] architecture in order to speed up convergence and achieve the highest level of accuracy.

**(Jike Zhong, Hong-You Chen et al. in 2023)** Research in [21] has shown that there may be a performance problem with training with BN in federated learning (FL) when non-IID decentralized data is used, as the BN statistics are mismatched between the training and testing phases. As a result, group normalization is more commonly used in FL as an alternative to BN. The purpose of this [21] paper is to identify a more fundamental problem with BN in FL that makes BN unsuitable even with high-frequency communication. In addition to this study, also the authors also reveal an unreasonable behavior of BN in FL. FL is relatively robust when communicating at low frequencies, where it is often assumed to degrade dramatically.

Furthermore, a great deal of progress has been made in the field of neural network description, with a large number of papers providing thorough explanations of training procedures, optimization strategies, and neural network topologies. Readers are invited to examine the works of [22-28] for a thorough overview of these subjects."

## 3. Proposed Work

In response to identified gaps and challenges in existing literature review our proposed work aims to address different questions raised in the batch normalization.

### Data set Description Preamble

Alzheimer's disease (A.D.) is a type of gradual neuro-disorder, a serious type of dementia very demanding to diagnose at an earlier stage without proper medical assistance. Alzheimer's disease cannot be cured at the moment; however, the disease can be managed under proper medical supervision. The onset of Alzheimer's disease symptoms normally appears gradually and worsens with time, interfering with daily tasks. Existent machine learning algorithms usually classify Alzheimer's disease (A.D.) based on a single input, an MRI scan of the brain. Neuro syndrome is a mental illness making it difficult for patients to remember previous events, lose their direction, interpret language, and make decisions. Patients as young as 65 can have Alzheimer's disease (A.D.), a serious form of dementia. The earliest detection of Alzheimer's disease is still difficult, even with scanned-based manual instruction and visual assessments. A set of datasets includes MRI images with a resolution of 256*256 in the axial plane. R. Polikar Rowan University provided an EEG dataset of an A.D. patient [29].

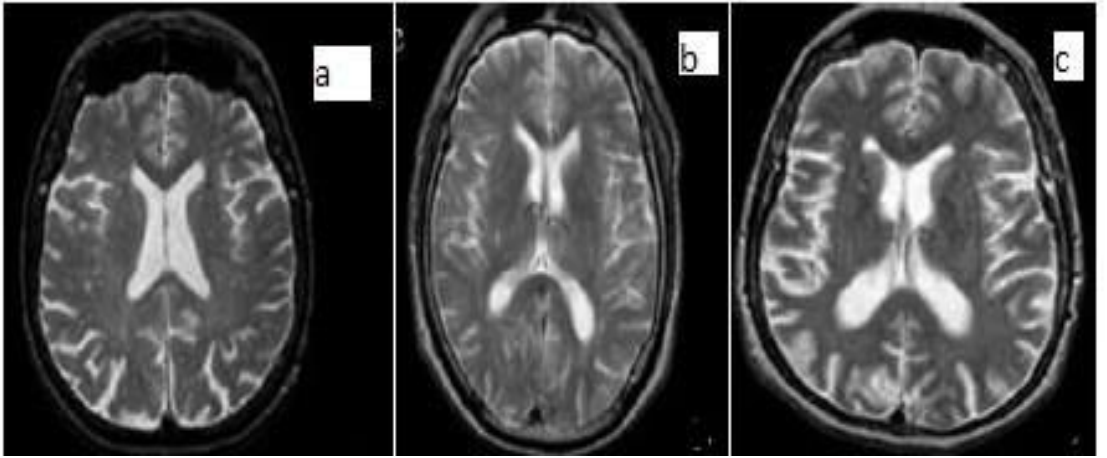

*Figure 1:* a) Normal brain images b) A.D. images; c) A.D. +visual agnosia.

**Hybrid classification model (Group normalized resnet 50) for detecting and categorizing Alzheimer's disease**.

In addition to IN and L.N., group normalization (G.N.) is another method that falls between them. Using the (H, W) axes across multiple channels, the system calculates I and I along groups of channels. The set of coefficients in the same input feature and the same group of channels as xi is called Si [30] [31]. The mathematical formulation behind group normalization is given as:

$$\mu_i = \frac{1}{n}\sum_{t \in S_i} y_t \qquad (6)$$

$$\sigma_i = \sqrt{\frac{1}{n}\sum_{t \in S_i}(y_t - \mu_i)^2 + \varepsilon} \qquad (7)$$

$$S_i = \left\{ t \mid t_m = i_m, \left\lfloor \frac{t_c}{C/G} \right\rfloor = \left\lfloor \frac{i_c}{C/G} \right\rfloor \right\} \qquad (8)$$

$$\hat{x}_i = \frac{1}{\sigma_i}(x_i - \mu_i) \qquad (9)$$

$$y_i = \gamma \hat{x}_i + \beta \qquad (10)$$

The algorithm for the proposed method is given as :

## Algorithm

**Input:**

  i. **S [ n,c, h, w], n: number of groups, c: channels, h: height of image, w: width of an image.**
  ii. **Other parameters** $(y, \alpha, \gamma, G, eps = 1e - \delta)$

**Output:**

*Classification results. of Alzhemer's disease.*

**Start**

**Step 1: def group normalization** $(y, \alpha, \gamma, G, eps = 1e - \delta)$:

  **# where y denotes input feature map with predefined shape as [ n,c, h, w]**

  **γ, β The scale and offset for G.N. are given as [1, C, 1, 1] and G represents different groups**

  Step 2: **n, c, h, w= y.shape**

  Step 3: **y = to.reshape (y, [ n ,G,c]**

  Step 4: **average mean, variance = to. No. Moments (y, [2, 3, 4], True dimensions should be maintained)**

  Step 5: **y = (y – mean) / tf. start (variance +eps)**

  Step 6: **y=tf. reshape( y, [n, c, h, w])**

**return y * γ +β**

**End.**

### Architectural details of the proposed model

ResNet-50 is a variant of ResNet, which stands for "Residual Network." ResNet was introduced to solve the problem of vanishing gradients associated with deep neural networks. As gradients propagate through multiple layers of a network, gradients disappear exponentially due to the chain rule of differentiation. The network is therefore unable to learn meaningful representations of the input data beyond a certain depth. This problem is addressed by ResNets skip connections, or shortcut connections that allow the network to learn residual mappings. Instead of learning the underlying mapping directly, the network learns the difference between inputs and outputs, which is called the residual. Adding the residual to the input produces the layer's output. By using this approach, the network can be trained deeper and learn deep representations more easily. The proposed architecture ResNet-50 is composed of 50 layers and comprises a convolutional layer followed by four stages containing multiple convolutional layers and skip connections. As the first stage of processing, the input feature maps are strided by 2, thereby reducing their spatial dimensions. In subsequent stages, the stride is 1, maintaining the spatial dimensions of the input feature maps. As the layers of a convolutional algorithm are increased, more filters are used to capture more complex features.

The ResNet-50 model has been widely used for image classification tasks, such as ImageNet, and has achieved state-of-the-art results. Aside from object detection, the architecture has also been adapted for semantic segmentation and semantic segmentation-related tasks in computer vision. A powerful CNN architecture known as ResNet-50 has helped advance the field of computer vision by allowing better and more accurate models to be developed. The proposed model's architecture can best be visualized by the graphical representation given in. figure 2. The hyper-parameters of the model are given in Table. 2

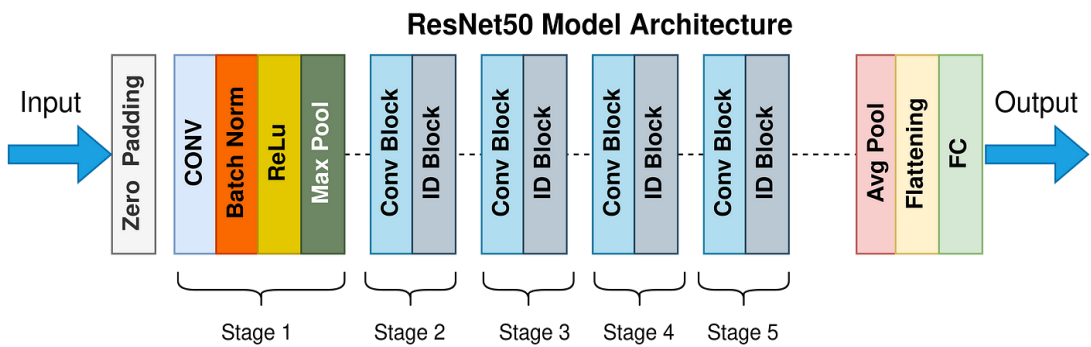

Figure 2. The architecture details of Resnet-50 Model

Table 2. Hyperparameters of the proposed model.

| Layer-name | 34-Layer | 50-layer | 101-layer |
|---|---|---|---|
| Conv1 | 7*7,64, stride 2 | | |
| | 3*3 max-pool, stride 2 | | |
| Conv2_x | [3*3,64  3*3,64] *3 | [3*3,64 3*3,64 1*1,256] *3 | [1*1,64  3*3,64 1*1,256] *3 |
| Conv3_x | [3*3,128  3*3,128] *4 | [1*1,128  3*3,128 1*1,512] *4 | [1*1,128  3*3,128 1*1,512] *4 |

| Conv4_x | [3*3,256  3*3,256] *6 | [1*1,256  3*3, 256 1*1,1024] *6 | [1*1,256  3*3, 256 1*1,1024] *6 |
| Conv5_x | [3*3,512 3*3,512] *3 | [1*1,512,3*3,512 1*1,2048] *3 | [1*1,512,3*3,512 1*1,2048] *3 |

### Performance matric

Evaluation of resnet-50's performance. The learning classifier is tested and compared to SVM, MLP, CNN, and DBN, among other techniques. It is divided into four types: Alzheimer's disease, mild Alzheimer's disease, Huntington's disease, and normal Alzheimer's disease. To accomplish this, we use the following performance evaluation metrics. $\text{TP}(\alpha_p), \text{TN}(\alpha_n)$, $\text{FP}(\beta_p)$ and $\text{FN}(\beta_n)$.

The Sensitivity Test (SEN): The Sensitivity Test (SEN) is a test that determines the likelihood of disease in a patient. It's also referred to as (Recall) or true positive rate (TPR).

A Specificity (SPE) is a measure of

$$\text{SPE} = \frac{\alpha_n}{\alpha_n + \beta_p} \quad (11)$$

The Accuracy (ACC):

$$\text{ACC} = \frac{\alpha_n + \alpha_p}{\alpha_n + \alpha_p + \beta_p + \beta_N} \quad (12)$$

MCC measures sensitivity and specificity in prediction by balancing sensitivity and specificity. The mathematics correlation coefficient (MCC) [25] can be used to construct accurate descriptions.

$$\text{MCC} = \frac{\alpha_n * \alpha_p - \beta_p * \beta_N}{\sqrt{(\alpha_p + \beta_p)*(\alpha_p + \beta_N)*(\alpha_n + \beta_N)}} \quad (13)$$

### Experiment Results

The presented model has been implemented on the medical dataset and performs the classification of Alzheimer's disease. Other models with batch normalization methods have been evaluated on the same dataset. Comparative analysis has been done in terms of accuracy, specificity, and MCC and also the comparison is done in terms of training loss, optimum training loss and variance. G.N. will be examined in the future on R.L. tasks, such as [31],[32], where B.N. is critical to training very deep models [33],[34]. After Testing resnet-50 on the given dataset, the experimental results obtained and compared with results of already existing techniques are recorded in Fig.3: On the Alzheimer's dataset, Figure 4 compares the proposed technique resnet-50 to the known method regarding training loss, training loss optimum, and variance. Graphs a, b and c in figure 3 determine how the training loss curve behaves in different models, the variance induced by each and the behaviour of the training loss optimum curve.

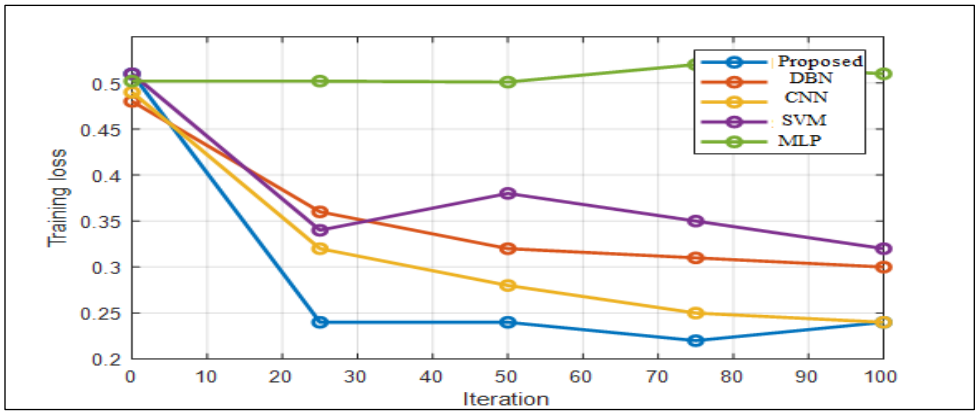

Figure 2. a) Training loss with varying iteration.

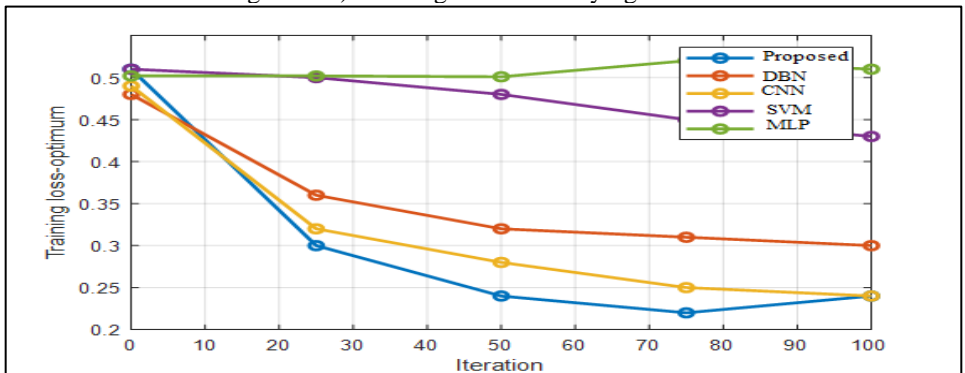

Figure 2. b) behaviour of Training loss optimum curve

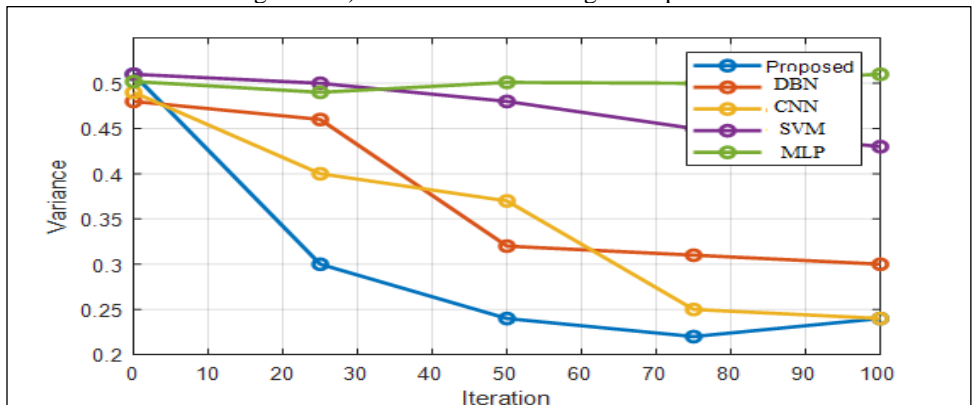

Figure2. c) variance curve behaviour

The performance metric comparison of the proposed technique with the other state-of-the-art techniques, such as in terms of ACC, SEN, SPE and MCC. The proposed technique achieves remarkable accuracy with the least error, thus outperforming all the techniques depicted in figure 3:

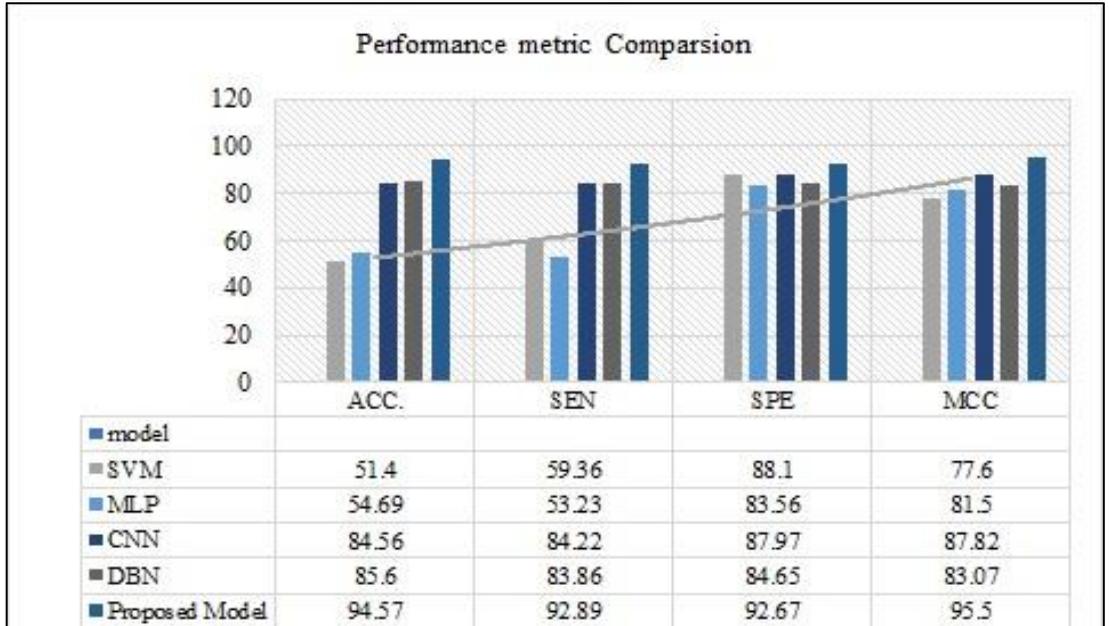

Figure3. Performance metric comparison

Table 2. Error rate comparison

| Input Dimensionality | Deep Learning Technique | Error rate (%) |
|---|---|---|
| 256*256 | SVM | 0.1256 |
| 256*256 | MLP | 0.1353 |
| 256*256 | CNN | 0.0245 |
| 256*256 | DBN | 0.0214 |
| 256*256 | Proposed technique | 0.0126 |

The proposed method achieves the least error rate compared to other SoTA techniques when tested against the same batch dimension depicted in table 2. In summary, our results show that our method built on top of G.N. successfully enhances the layer's capability to help train the neural network through modification of the underlying mechanism. In this way, the G.N. layer becomes more robust to hyperparameter changes. Consequently, it is more likely to prevent gradient vanishing or explosion problems.

## 4. Conclusion

To conclude, normalization plays a critical role in enhancing the performance, stability, and generalization abilities of machine learning models. A number of normalization techniques, including batch normalization, layer normalization, instance normalization, and group normalization, can help mitigate the impact of internal covariate shifts and prevent overfitting, which are common challenges in deep learning and other complex models. Building high-quality models that solve complex problems in a wide range of domains requires understanding and applying normalization techniques. We can enhance the robustness, accuracy, and efficiency of our model's using normalization, making it a valuable tool for modern machine learning practitioners. While batch normalization has been a popular method of normalization for deep learning, its limitations can make it challenging for certain applications, such as object detection and semantic segmentation.

### Limitations and Future Scope

There are limitations to this approach, including a need for large batch sizes to approximate population mean and variance, as well as the fact that the fixed mean and variance derived from the training data may not be generalizable to the test data. A promising alternative to batch normalization (B.N.) is group normalization (G.N.) which divides features into groups to normalize them. G.N can perform better than other normalization techniques in deep learning models by tuning the hyperparameter num_groups. It is a useful alternative to B.N for deep learning practitioners dealing with high-resolution images and videos that require lots of memory. There have been a variety of standardization layers proposed for training neural networks. The Group Normalization (G.N.) study achieved significant performance in the visual recognition task through its effective and attractive methodology. We proposed the Resnet50 model and presented a group normalization method that is the most effective normalizing layer without relying on batch dimensions in this paper.

Though G.N. achieves better performance than the B.N. layer, it still leads to gradient explosion under certain conditions. The problem can be resolved by introducing extra regularization terms into the parameter $\lambda$. Our future work will include testing the proposed layer in larger datasets (e.g., ImageNet) and applying it to other vision tasks, such as object detection and video classification. Moreover, we will also learn other theoretical studies to better understand the G.N. layer's effect on network optimization, thus improving G.N.

### Conflicts of Interest

The authors declare that they have no competing interests.

### References

[1] Luo, P., Zhanglin, P., Wenqi, S., Ruimao, Z., Jiamin, R., & Lingyun, W. (2019, May). Differentiable dynamic normalization for learning deep representation, In *International Conference on Machine Learning* (pp. 4203-4211). PMLR.

[2] I. Ioffe S, Szegedy C (2015) Batch normalization: accelerating deep network training by reducing internal covariate shift.

[3]. Garbin, C., Zhu, X., & Marques, O. (2020). Dropout vs batch normalization: an empirical study of their impact on deep learning. *Multimedia Tools and Applications*, 1-39.

[4]. Gao, S. H., Han, Q., Li, D., Cheng, M. M., & Peng, P. (2021). Representative batch normalization with feature calibration. In *Proceedings of the IEEE/CVF Conference on Computer Vision and Pattern Recognition* (pp. 8669-8679).

[5]. Xiong, R., Yang, Y., He, D., Zheng, K., Zheng, S., Xing, C., ... & Liu, T. (2020, November). On-layer normalization in the transformer architecture. In *International Conference on Machine Learning* (pp. 10524-10533). PMLR.

[6]. Liu, F., Ren, X., Zhang, Z., Sun, X., & Zou, Y. (2020, December). Rethinking skip connection with layer normalization. In *Proceedings of the 28th International Conference on Computational Linguistics* (pp. 3586-3598).

[7]. Jing, Y., Liu, X., Ding, Y., Wang, X., Ding, E., Song, M., & Wen, S. (2020, April). Dynamic instance normalization for arbitrary style transfer. In *Proceedings of the AAAI Conference on Artificial* Intelligence (Vol. 34, No. 04, pp. 4369-4376).

[8]. Syed Waqas Zamir, Aditya Arora, Salman Khan, Munawar Hayat, Fahad Shahbaz Khan, Ming-Hsuan Yang, and Ling Shao. Multi-stage progressive image restoration. arXiv preprint arXiv:2102.02808, 2021.

[9]. Seungjun Nah, Sanghyun Son, Sooyoung Lee, Radu Timofte, Kyoung Mu Lee, et al. NTIRE 2021 challenge on image deblurring. In IEEE/CVF Conference on Computer Vision and Pattern Recognition Workshops, 2021.

[10]. Chen, L., Lu, X., Zhang, J., Chu, X., & Chen, C. (2021). Hemet: Half instance normalization network for image restoration. In Proceedings of the IEEE/CVF Conference on Computer Vision and Pattern Recognition (pp. 182-192).

[11]. Lepcha, D. C., Goyal, B., & Dogra, A. (2020). Image Fusion based on Cross Bilateral and Rolling Guidance Filter through Weight Normalization. *The Open Neuroimaging Journal*, *13*(1).

[12]. Yuan, X., Shen, X., Mehta, S., Li, T., Ge, S., & Zha, Z. (2021). Structure injected weight normalization for training deep networks. *Multimedia Systems*, 1-12.

[13]. Luo, P., Zhang, R., Ren, J., Peng, Z., & Li, J. (2019). Switchable normalization for learning-to-normalize deep representation. *IEEE transactions on pattern analysis and machine intelligence*, *43*(2), 712-728.

[14]. Neupane, D., Kim, Y., & Seok, J. (2021). Bearing Fault Detection Using Scalogram and Switchable Normalization-Based CNN (SN-CNN). *IEEE Access*.

[15]. Zhou, K., Huang, X., Li, Y., Zha, D., Chen, R., & Hu, X. (2020). Towards deeper graph neural networks with differentiable group normalization. *arXiv preprint arXiv:2006.06972*.

[16]. Zhou, X. Y., Sun, J., Ye, N., Lan, X., Luo, Q., Lai, B. L., ... & Li, Z. (2020). Batch Group Normalization. *arXiv preprint arXiv:2012.02782*.

[17]. Gunawan, A., Yin, X., & Zhang, K. (2022). Understanding and Improving Group Normalization. arXiv preprint arXiv:2207.01972.

[18]. Sinha, D., & Thangavel, K. (2022). Automatic epileptic signal classification using a deep convolutional neural network. Journal of Discrete Mathematical Sciences and Cryptography, 25(4), 963-973.

[19]. Kumar, R., & Kumar, D. (2022). Comparative analysis of validating parameters in the deep learning models for remotely sensed images. Journal of Discrete Mathematical Sciences and Cryptography, 25(4), 913-920.

[20]. Gilbert Chandra, D., Srinivasulu Reddy, U., Uma, G., & Umapathy, M. (2023). Group normalization-based 2D-convolutional neural network for intelligent bearing fault diagnosis. Journal of the Brazilian Society of Mechanical Sciences and Engineering, 45(11), 584.

[21]. Casella, B., Esposito, R., Sciarappa, A., Cavazzoni, C., & Aldinucci, M. (2023). Experimenting with Normalization Layers in Federated Learning on non-IID scenarios. arXiv preprint arXiv:2303.10630.

[22]. Cao, Y., Chandrasekar, A., Radhika, T., & Vijayakumar, V. (2023). Input-to-state stability of stochastic Markovian jump genetic regulatory networks. Mathematics and Computers in Simulation.

[23]. Radhika, T., Chandrasekar, A., Vijayakumar, V., & Zhu, Q. (2023). Analysis of Markovian jump stochastic Cohen–Grossberg BAM neural networks with time delays for exponential input-to-state stability. Neural Processing Letters, 55(8), 11055-11072.

[24]. Rakkiyappan, R., Premalatha, S., Chandrasekar, A., & Cao, J. (2016). Stability and synchronization analysis of inertial memristive neural networks with time delays. Cognitive neurodynamics, 10, 437-451.

[25]. Habib, G., & Qureshi, S. (2023). Compressed lightweight deep learning models for resource-constrained Internet of things devices in the healthcare sector. Expert Systems, e13269.

[26]. Habib, G., & Qureshi, S. (2022). GAPCNN with HyPar: Global Average Pooling convolutional neural network with novel NNLU activation function and HYBRID parallelism. Frontiers in Computational Neuroscience, 16, 1004988.

[27]. Habib, G., & Qureshi, S. (2021). Comparative analysis of lbp variants with the introduction of new radial and circumferential derivatives. International Journal of Computing and Digital System.

[28]. Habib, G., & Qureshi, S. (2020). Biomedical image classification using CNN by exploiting deep domain transfer learning. International Journal of Computing and Digital Systems, 10, 2-11.

[29]. https://adni.loni.ucla.edu.

[30]. https://sh-tsang.medium.com/review-group-norm-gn-group-normalization-image-classification.

[31]. Wu, Y., & He, K. (2018). Group normalization. In *Proceedings of the European conference on computer vision (ECCV)* (pp. 3-19).

[32]. M.M. Dessouky, M.A. Elrashidy, T.E. Taha, and H.M. Abdelkader, "Selecting and Extracting Effective Features for Automated Diagnosis of Alzheimer's Disease", International Journal of Computer Applications, Vol. 81 – No.4, pp 17-28, (2013).

[33]. D. Silver, J. Schrittwieser, K. Simonyan, I. Antonoglou, A. Huang, A. Guez, T. Hubert, L. Baker, M. Lai, A. Bolton, Y. Chen, T. Lillicrap, F. Hui, L. Sifre, G. van den Driessche, T. Graepel, and D. Hassabis. Mastering the game of go without human knowledge. Nature, 2017.

[34]. K. He, X. Zhang, S. Ren, and J. Sun. Deep residual learning for image recognition. In CVPR, 2016. H.H. Crokell. "Specialization and International Competitiveness", *Managing the Multinational Subsidiary,* H. Etemad and L. S, Sulude (eds.), Croom-Helm, London, 1986.